\documentclass[10pt,onecolumn,draftclsnofoot]{IEEEtran}

\usepackage[square,numbers,sort&compress]{natbib}

\ifCLASSINFOpdf
\else
\fi
\usepackage{epsfig}
\usepackage{graphicx}
\usepackage{subfigure}
\usepackage[cmex10]{amsmath}
\usepackage{mathrsfs}
\usepackage{amsmath}

\usepackage{algorithm}

\begin{document}
\title{Tensor Representation and Manifold Learning Methods for Remote Sensing Images}

\author{{Lefei~Zhang,~\IEEEmembership{Member,~IEEE}} % <-this % stops a space

\thanks{Date of Current Version: January 12, 2014.}
\thanks{L. Zhang is with the Computer School, Wuhan University, Wuhan 430072, China (e-mail: zhanglefei@whu.edu.cn)}% <-this % stops a space
\thanks{This PH. D. Thesis is supervised by Prof. Liangpei Zhang and Prof. Dacheng Tao.}}% <-this % stops a space

\markboth{PH. D. Thesis of Wuhan University}%
{Zhang \MakeLowercase{\textit{et al.}}: }

\maketitle

\begin{abstract}
  One of the main purposes of earth observation is to extract interested information and knowledge from remote sensing (RS) images with high efficiency and accuracy. However, with the development of RS technologies, RS system provide images with higher spatial and temporal resolution and more spectral channels than before, and it is inefficient and almost impossible to manually interpret these images. Thus, it is of great interests to explore automatic and intelligent algorithms to quickly process such massive RS data with high accuracy. This thesis targets to develop some efficient information extraction algorithms for RS images, by relying on the advanced technologies in machine learning. More precisely, we adopt the manifold learning algorithms as the mainline and unify the regularization theory, tensor-based method, sparse learning and transfer learning into the same framework. The main contributions of this thesis are as follows.
\end{abstract}

\IEEEpeerreviewmaketitle

\subsection{Chapter I}
  \IEEEPARstart{W}{e} introduced a patch alignment framework to unify some conventional dimension reduction (DR) methods (e.g. PCA and LDA) and some current and representative manifold learning algorithms (e.g. LLE, ISOMAP, LE, LTSA and HLLE) into the same optimization framework. This framework reveals the basic principle of manifold learning algorithms: 1) algorithms are intrinsically different in the patch optimization stage, and 2) all algorithms share an almost identical alignment stage. The framework also enables us to develop new dimensionality reduction algorithms by modify the local alignment matrix for new applications.

  \subsubsection{LPDML}
  In hyperspectral image (HSI) classification, feature extraction is one important step. Traditional methods, e.g., principal component analysis (PCA) and locality preserving projection, usually neglect the information of within-class similarity and between-class dissimilarity, which is helpful to the improvement of classification. On the other hand, most of these methods, e.g., PCA and linear discriminative analysis, consider that the HSI data lie on a low-dimensional manifold or each class is on a submanifold. However, some class data of HSI may lie on a multimanifold. To avoid these problems, we propose a method for feature extraction in HSIs, assuming that a local region resides on a submainfold. In our method, we deal with the data region by region by taking into account the different discriminative locality information. Then, under the metric learning framework, a robust distance metric is learned. It aims to learn a subspace in which the samples in the same class are as near as possible while the samples in different classes are as far as possible. Encouraging experimental results on two available hyperspectral data sets indicate that our proposed algorithm outperforms many existing feature extract methods for HSI classification \cite{LPDML}.

  \subsubsection{LDLE}
  In this paper, the linear discriminative Laplacian eigenmaps (LDLE) dimensionality reduction (DR) algorithm is introduced to C-band polarimetric synthetic aperture radar (PolSAR) agricultural classification. A collection of homogenous areas of the same crop class usually presents physical parameter variation, such as the biomass and soil moisture. Furthermore, the local incidence angle also impacts a lot on the same crop category when the vegetation layer is penetrable with C-band radar. We name this phenomenon as the ¡°observed variation of the same category¡± (OVSC). The most common PolSAR features, e.g., the Freeman¨CDurden and Cloude¨CPottier decompositions, show an inadequate performance with OVSC. In our research, more than 40 coherent and incoherent PolSAR decomposition models are stacked into the high-dimensionality feature cube to describe the various physical parameters. The LDLE algorithm is then performed on the observed feature cube, with the aim of simultaneously pushing the local samples of the same category closer to each other, as well as maximizing the distance between local samples of different categories in the learnt subspace. Finally, the classification result is obtained by nearest neighbor (NN) or Wishart classification in the reduced feature space. In the simulation experiment, eight crop blocks are picked to generate a test patch from the 1991 Airborne Synthetic Aperture Radar (AIRSAR) C-band fully polarimetric data from of Flevoland test site. Locality preserving projections (LPP) and principal component analysis (PCA) are then utilized to evaluate the DR results of the proposed method. The classification results show that LDLE can distinguish the influence of the physical parameters and achieve a 99\% overall accuracy, which is better than LPP (97\%), PCA (88\%), NN (89\%), and Wishart (88\%). In the real data experiment, the Chinese Hailaer nationalized farm RadarSat2 PolSAR test set is used, and the classification accuracy is around 94\%, which is again better than LPP (90\%), PCA (88\%), NN (89\%), and Wishart (85\%). Both experiments suggest that the LDLE algorithm is an effective way of relieving the OVSC phenomenon \cite{LDLE}.

  \subsubsection{NDML}
  Manifold learning algorithms have been demonstrated to be effective for hyperspectral data dimension reduction (DR). However, the low dimensional feature representation resulted by traditional manifold learning algorithms could not preserve the nonnegative property of the hyperspectral data, which leads inconsistency with the psychological intuition of ¡°combining parts to form a whole¡±. In this paper, we introduce a nonnegative discriminative manifold learning (NDML) algorithm for hyperspectral data DR, which yields a discriminative and low dimensional feature representation, with psychological and physical evidence in the human brain. Our method benefits from both the nonnegative matrix factorization (NMF) algorithm and the discriminative manifold learning (DML) algorithm. We apply the NDML algorithm to hyperspectral remote sensing image classification on HYDICE dataset. Experimental results confirm the efficiency of the proposed NDML algorithm, compared with some existing manifold learning based DR methods \cite{NDML}.

  \subsubsection{SGE}
  This letter introduces an efficiency-manifold-learning-based supervised graph embedding (SGE) algorithm for polarimetric synthetic aperture radar (POLSAR) image classification. We use a linear dimensionality reduction technology named SGE to obtain a low-dimensional subspace which can preserve the discriminative information from training samples. Various POLSAR decomposition features are stacked into the input feature cube in the original high-dimensional feature space. The SGE is then implemented to project the input feature into the learned subspace for subsequent classification. The suggested method is validated by the full polarimetric airborne SAR system EMISAR, in Foulum, Denmark. The experiments show that the SGE presents a favorable classification accuracy and the valid components of the multifeature cube are also distinguished \cite{SGE}.

  \subsubsection{ADML}
  Manifold learning methods have widely used in ordinary image processing domain. It has many advantages, depending on the different formulation of the manifold. Hyperspectral images are kind of images acquired by air-borne or space-born platforms. This paper introduces a novel manifold learning based dimension reduction (DR) method for hyperspectral classification. The purpose is to fully utilize the spectral and spatial information from hyperspectral images to get confidential landcover and land use class results \cite{ADML}.

\subsection{Chapter II}
  We extensively discussed the tensor-based manifold learning methods in RS image analysis. We also propose a method for the dimensionality reduction of spectral-spatial features in RS images, under the umbrella of multi-linear algebra, i.e., the algebra of tensors. The proposed approach is a tensor extension of conventional supervised manifold learning based DR. In particular, we define a tensor organization scheme for representing a pixel¡¯s spectral-spatial feature and develop tensor discriminative locality alignment (TDLA) for removing redundant information for subsequent classification.

  \subsubsection{TDLA}
  In this paper, we propose a method for the dimensionality reduction (DR) of spectral-spatial features in hyperspectral images (HSIs), under the umbrella of multilinear algebra, i.e., the algebra of tensors. The proposed approach is a tensor extension
  of conventional supervised manifold-learning-based DR. In particular, we define a tensor organization scheme for representing a pixel¡¯s spectral-spatial feature and develop tensor discriminative locality alignment (TDLA) for removing redundant information for subsequent classification. The optimal solution of TDLA is obtained by alternately optimizing each mode of the input tensors. The methods are tested on three public real HSI data sets collected by hyperspectral digital imagery collection experiment, reflective optics system imaging spectrometer, and airborne visible/infrared imaging spectrometer. The classification results show significant improvements in classification accuracies while using a small number of features \cite{TDLA}.

\subsection{Chapter III}
  We proposed two multiple feature combining (MFC) algorithms to deal with RS image classification tasks with multiple features as input. In these two methods, we first use the LE and SNE to build the local alignment matrix, respectively. Then, the proposed adaptive manifold learning MFC algorithms combine the input multiple features linearly in an optimal way and obtain a unified low-dimensional representation of these multiple features for subsequent classification. Each feature has its particular contribution to the unified representation determined by simultaneously optimizing the weights in the objective function.

  \subsubsection{MFC}
  In hyperspectral remote sensing image classification, multiple features, e.g., spectral, texture, and shape features, are employed to represent pixels from different perspectives. It has been widely acknowledged that properly combining multiple features always results in good classification performance. In this paper, we introduce the patch alignment framework to linearly combine multiple features in the optimal way and obtain a unified low-dimensional representation of these multiple features for subsequent classification. Each feature has its particular contribution to the unified representation determined by simultaneously optimizing the weights in the objective function. This scheme considers the specific statistical properties of each feature to achieve a physically meaningful unified low-dimensional representation of multiple features. Experiments on the classification of the hyperspectral digital imagery collection experiment and reflective optics system imaging spectrometer hyperspectral data sets suggest that this scheme is effective \cite{MFC}.

  \subsubsection{MSNE}
  In automated remote sensing based image analysis, it is important to consider the multiple features of a certain pixel, such as the spectral signature, morphological property, and shape feature, in both the spatial and spectral domains, to improve the classification accuracy. Therefore, it is essential to consider the complementary properties of the different features and combine them in order to obtain an accurate classification rate. In this paper, we introduce a modified stochastic neighbor embedding (MSNE) algorithm for multiple features dimension reduction (DR) under a probability preserving projection framework. For each feature, a probability distribution is constructed based on t-distributed stochastic neighbor embedding (t-SNE), and we then alternately solve t-SNE and learn the optimal combination coefficients for different features in the proposed multiple features DR optimization. Compared with conventional remote sensing image DR strategies, the suggested algorithm utilizes both the spatial and spectral features of a pixel to achieve a physically meaningful low-dimensional feature representation for the subsequent classification, by automatically learning a combination coefficient for each feature. The classification results using hyperspectral remote sensing images (HSI) show that MSNE can effectively improve RS image classification performance \cite{MSNE}.

\subsection{Chapter IV}
  We proposed two regularized discriminative manifold learning algorithms for the detection of hyper-spectral targets, namely sparse transfer manifold embedding (STME) and supervised metric learning (SML). Technically speaking, these methods are particularly designed for hyperspectral target detection by introducing the multiple constraints into discriminative manifold learning framework. In the STME, a sparse formulation and transfer regularization are adopted, while in the SML, a similarity propagation constraint and a smoothness regularization on manifold are imposed. Both of the algorithms have showed their outstanding performance in targets detection.

  \subsubsection{STME}
  Target detection is one of the most important applications in hyperspectral remote sensing image analysis. However, the state-of-the-art machine-learning-based algorithms for hyperspectral target detection cannot perform well when the training samples, especially for the target samples, are limited in number. This is because the training data and test data are drawn from different distributions in practice and given a small-size training set in a high-dimensional space, traditional learning models without the sparse constraint face the over-fitting problem. Therefore, in this paper, we introduce a novel feature extraction algorithm named sparse transfer manifold embedding (STME), which can effectively and efficiently encode the discriminative information from limited training data and the sample distribution information from unlimited test data to find a low-dimensional feature embedding by a sparse transformation. Technically speaking, STME is particularly designed for hyperspectral target detection by introducing sparse and transfer constraints. As a result of this, it can avoid over-fitting when only very few training samples are provided. The proposed feature extraction algorithm was applied to extensive experiments to detect targets of interest, and STME showed the outstanding detection performance on most of the hyperspectral datasets \cite{STME}.

  \subsubsection{SML}
  The detection and identification of target pixels such as certain minerals and man-made objects from hyperspectral remote sensing images is of great interest for both civilian and military applications. However, due to the restriction in the spatial resolution of most airborne or satellite hyperspectral sensors, the targets often appear as subpixels in the hyperspectral image (HSI). The observed spectral feature of the desired target pixel (positive sample) is therefore a mixed signature of the reference target spectrum and the background pixels spectra (negative samples), which belong to various land cover classes. In this paper, we propose a novel supervised metric learning (SML) algorithm, which can effectively learn a distance metric for hyperspectral target detection, by which target pixels are easily detected in positive space while the background pixels are pushed into negative space as far as possible. The proposed SML algorithm first maximizes the distance between the positive and negative samples by an objective function of the supervised distance maximization. Then, by considering the variety of the background spectral features, we put a similarity propagation constraint into the SML to simultaneously link the target pixels with positive samples, as well as the background pixels with negative samples, which helps to reject false alarms in the target detection. Finally, a manifold smoothness regularization is imposed on the positive samples to preserve their local geometry in the obtained metric. Based on the public data sets of mineral detection in an Airborne Visible/Infrared Imaging Spectrometer image and fabric and vehicle detection in a Hyperspectral Mapper image, quantitative comparisons of several HSI target detection methods, as well as some state-of-the-art metric learning algorithms, were performed. All the experimental results demonstrate the effectiveness of the proposed SML algorithm for hyperspectral target detection \cite{SML}.

\subsection{Chapter V}
  We investigated a general multi-linear data analysis framework for tensor inputs for RS image classification. This framework generalizes the current classifiers which only accept vectors as inputs into multi-linear condition. Based on this framework, we further proposed some algorithms, e.g. support tensor machine, multiclass support tensor machine, and proximal support tensor machine. Experiments and comparisons show the effectiveness of tensor representation and analysis approaches in RS image information extraction with a small number of training samples.

  \subsubsection{STM}
  In remote-sensing image target recognition, the target or background object is usually transformed to a feature vector, such as a spectral feature vector. However, this kind of vector represents only one pixel of a remote-sensing image that considers the spectral information but ignores the spatial relationship of neighboring pixels (i.e., the local texture and structure). In this letter, we propose a new way to represent an image object as a multifeature tensor that encodes both the spectral and textural information (Gabor function) and then apply the support tensor machine for target recognition. A range of experiments demonstrates that the effectiveness of the proposed method can deliver a high and correct recognition rate with a small number of training samples \cite{STM}.

  \subsubsection{R1TD}
  In this study, a novel noise reduction algorithm for hyperspectral imagery (HSI) is proposed based on high-order rank-1 tensor decomposition. The hyperspectral data cube is considered as a three-order tensor that is able to jointly treat both the spatial and spectral modes. Subsequently, the rank-1 tensor decomposition (R1TD) algorithm is applied to the tensor data, which takes into account both the spatial and spectral information of the hyperspectral data cube. A noise-reduced hyperspectral image is then obtained by combining the rank-1 tensors using an eigenvalue intensity sorting and reconstruction technique. Compared with the existing noise reduction methods such as the conventional channel-by-channel approaches and the recently developed multidimensional filter, the spatial¨Cspectral adaptive total variation filter, experiments with both synthetic noisy data and real HSI data reveal that the proposed R1TD algorithm significantly improves the HSI data quality in terms of both visual inspection and image quality indices. The subsequent image classification results further validate the effectiveness of the proposed HSI noise reduction algorithm \cite{R1TD}.

\section*{Acknowledgment}
  I would like to dedicate this thesis to my mother, in memory of her love forever. I owe my family a lot for many things. I also thank all my family members for their extreme understanding and infinite support of my PhD study and academic pursuits.

\ifCLASSOPTIONcaptionsoff
  \newpage
\fi

\bibliographystyle{IEEEtran}
\footnotesize
\bibliography{Ref}

\begin{IEEEbiographynophoto}{Lefei Zhang}
(S'11$-$M'14) received the B.S. degree in sciences and techniques of remote sensing and the Ph.D. degree in photogrammetry and remote sensing from Wuhan University, Wuhan, China, in 2008 and 2013, respectively.

In August 2013, He joined the the Computer School, Wuhan University, where he is currently an Assistant Professor. His research interests include hyperspectral data analysis, high-resolution image processing, and pattern recognition in remote sensing images.

Dr. Zhang is a Reviewer of more than 10 international journals, including the \textsc{IEEE Transactions on Geoscience and Remote Sensing}, \textsc{IEEE Journal of Selected Topics in Applied Earth Observations and Remote Sensing}, \textsc{IEEE Geoscience and Remote Sensing Letters}, \emph{Information Sciences}, and \emph{Pattern Recognition}.
\end{IEEEbiographynophoto}

\vfill

\end{document}